\documentclass[a4paper,twoside]{article}

\usepackage{epsfig}
\usepackage{subcaption}
\usepackage{calc}
\usepackage{amssymb}
\usepackage{amstext}
\usepackage{amsmath}
\usepackage{amsthm}
\usepackage{multicol}
\usepackage{pslatex}
\usepackage{apalike}
\usepackage[bottom]{footmisc}
\usepackage[nolist,nohyperlinks]{acronym}

\usepackage{graphicx}
\usepackage{amsfonts}

\usepackage{standalone}

\usepackage{graphicx}
\usepackage{color}
\usepackage{multirow}
\usepackage{etoolbox}
\usepackage{textcomp}

\graphicspath{{./images/}}

\usepackage{xcolor}
\usepackage{tikz}
\usepackage{graphicx}
\usepackage{pgfplots}
\usetikzlibrary{spy}
\usetikzlibrary{lindenmayersystems}
\usetikzlibrary[shadings]
\usepgfplotslibrary{groupplots}

\usepackage{multirow}

\usepackage{algorithm}
\usepackage{algpseudocode}

\algnewcommand\algorithmicforeach{\textbf{for each}}
\algdef{S}[FOR]{ForEach}[1]{\algorithmicforeach\ #1\ \algorithmicdo}

\usepackage{url}
\usepackage[bookmarks=false]{hyperref}

\usepackage{SCITEPRESS}     

\newcommand\copyrighttext{%
  \footnotesize 
  \textcopyright This paper was accepted for presentation at the ICAART 2023 : 15th International Conference on Agents and Artificial Intelligence, Lisbon, Portugal.
  }
\newcommand\copyrightnotice{%
\begin{tikzpicture}[remember picture,overlay]
\node[anchor=north,yshift=-10pt] at (current page.north) {\fbox{\parbox{\dimexpr\textwidth-\fboxsep-\fboxrule\relax}{\copyrighttext}}};
\end{tikzpicture}%
}

\begin{document}

\title{Deep W-Networks: Solving Multi-Objective Optimisation Problems With Deep Reinforcement Learning}

\author{\authorname{Jernej Hribar, Luke Hackett, and Ivana Dusparic}
\affiliation{School of Computer Science and Statistics, Trinity College Dublin, Ireland}
\email{\{jhribar, lhackett, Ivana.Dusparic\}@tcd.ie}
}

\keywords{Deep Reinforcement Learning, Deep Q-Networks, W-Learning, Deep W-Networks, and Multi-objective.}

\begin{acronym}[MACHU]
  \acro{iot}[IoT]{Internet of Things}
  \acro{cr}[CR]{Cognitive Radio}
  \acro{ofdm}[OFDM]{orthogonal frequency-division multiplexing}
  \acro{ofdma}[OFDMA]{orthogonal frequency-division multiple access}
  \acro{scfdma}[SC-FDMA]{single carrier frequency division multiple access}
  \acro{rbi}[RBI]{ Research Brazil Ireland}
  \acro{rfic}[RFIC]{radio frequency integrated circuit}
  \acro{sdr}[SDR]{Software Defined Radio}
  \acro{sdn}[SDN]{Software Defined Networking}
  \acro{su}[SU]{Secondary User}
  \acro{ra}[RA]{Resource Allocation}
  \acro{qos}[QoS]{quality of service}
  \acro{usrp}[USRP]{Universal Software Radio Peripheral}
  \acro{mno}[MNO]{Mobile Network Operator}
  \acro{mnos}[MNOs]{Mobile Network Operators}
  \acro{gsm}[GSM]{Global System for Mobile communications}
  \acro{tdma}[TDMA]{Time-Division Multiple Access}
  \acro{fdma}[FDMA]{Frequency-Division Multiple Access}
  \acro{gprs}[GPRS]{General Packet Radio Service}
  \acro{msc}[MSC]{Mobile Switching Centre}
  \acro{bsc}[BSC]{Base Station Controller}
  \acro{umts}[UMTS]{universal mobile telecommunications system}
  \acro{Wcdma}[WCDMA]{Wide-band code division multiple access}
  \acro{wcdma}[WCDMA]{wide-band code division multiple access}
  \acro{cdma}[CDMA]{code division multiple access}
  \acro{lte}[LTE]{Long Term Evolution}
  \acro{papr}[PAPR]{peak-to-average power rating}
  \acro{hn}[HetNet]{heterogeneous networks}
  \acro{phy}[PHY]{physical layer}
  \acro{mac}[MAC]{medium access control}
  \acro{amc}[AMC]{adaptive modulation and coding}
  \acro{mimo}[MIMO]{multiple input multiple output}
  \acro{rats}[RATs]{radio access technologies}
  \acro{vni}[VNI]{visual networking index}
  \acro{rbs}[RB]{resource blocks}
  \acro{rb}[RB]{resource block}
  \acro{ue}[UE]{user equipment}
  \acro{cqi}[CQI]{Channel Quality Indicator}
  \acro{hd}[HD]{half-duplex}
  \acro{fd}[FD]{full-duplex}
  \acro{sic}[SIC]{self-interference cancellation}
  \acro{si}[SI]{self-interference}
  \acro{bs}[BS]{base station}
  \acro{fbmc}[FBMC]{Filter Bank Multi-Carrier}
  \acro{ufmc}[UFMC]{Universal Filtered Multi-Carrier}
  \acro{scm}[SCM]{Single Carrier Modulation}
  \acro{isi}[ISI]{inter-symbol interference}
  \acro{ftn}[FTN]{Faster-Than-Nyquist}
  \acro{m2m}[M2M]{machine-to-machine}
  \acro{mtc}[MTC]{machine type communication}
  \acro{mmw}[mmWave]{millimeter wave}
  \acro{bf}[BF]{beamforming}
  \acro{los}[LOS]{line-of-sight}
  \acro{nlos}[NLOS]{non line-of-sight}
  \acro{capex}[CAPEX]{capital expenditure}
  \acro{opex}[OPEX]{operational expenditure}
  \acro{ict}[ICT]{information and communications technology}
  \acro{sp}[SP]{service providers}
  \acro{inp}[InP]{infrastructure providers}
  \acro{mvnp}[MVNP]{mobile virtual network provider}
  \acro{mvno}[MVNO]{mobile virtual network operator}
  \acro{nfv}[NFV]{network function virtualization}
  \acro{vnfs}[VNF]{virtual network functions}
  \acro{cran}[C-RAN]{Cloud Radio Access Network}
  \acro{bbu}[BBU]{baseband unit}
  \acro{bbus}[BBU]{baseband units}
  \acro{rrh}[RRH]{remote radio head}
  \acro{rrhs}[RRH]{Remote radio heads} 
  \acro{sfv}[SFV]{sensor function virtualization}
  \acro{wsn}[WSN]{wireless sensor networks} 
  \acro{bio}[BIO]{Bristol is open}
  \acro{vitro}[VITRO]{Virtualized dIstributed plaTfoRms of smart Objects}
  \acro{os}[OS]{operating system}
  \acro{www}[WWW]{world wide web}
  \acro{iotvn}[IoT-VN]{IoT virtual network}
  \acro{mems}[MEMS]{micro electro mechanical system}
  \acro{mec}[MEC]{Mobile edge computing}
  \acro{coap}[CoAP]{Constrained Application Protocol}
  \acro{vsn}[VSN]{Virtual sensor network}
  \acro{rest}[REST]{REpresentational State Transfer}
  \acro{aoi}[AoI]{Age of Information}
  \acro{lora}[LoRa\texttrademark]{Long Range}
  \acro{iot}[IoT]{Internet of Things}
  \acro{snr}[SNR]{Signal-to-Noise Ratio}
  \acro{cps}[CPS]{Cyber-Physical System}
  \acro{uav}[UAV]{Unmanned Aerial Vehicle}
  \acro{rfid}[RFID]{Radio-frequency identification}
  \acro{lpwan}[LPWAN]{Low-Power Wide-Area Network}
  \acro{lgfs}[LGFS]{Last Generated First Served}
  \acro{wsn}[WSN]{wireless sensor network} 
  \acro{lmmse}[LMMSE]{Linear Minimum Mean Square Error}
  \acro{rl}[RL]{Reinforcement Learning}
  \acro{nb-iot}[NB-IoT]{Narrowband IoT}
  \acro{lorawan}[LoRaWAN]{Long Range Wide Area Network}
  \acro{mdp}[MDP]{Markov Decision Process}
  \acro{ann}[ANN]{Artificial Neural Network}
  \acro{dqn}[DQN]{Deep Q-Networks}
  \acro{mse}[MSE]{Mean Square Error}
  \acro{ml}[ML]{Machine Learning}
  \acro{cpu}[CPU]{Central Processing Unit}
  \acro{ddpg}[DDPG]{Deep Deterministic Policy Gradient}
  \acro{ai}[AI]{Artificial Intelligence}
  \acro{gp}[GP]{Gaussian Processes}
  \acro{drl}[DRL]{Deep Reinforcement Learning}
  \acro{mmse}[MMSE]{Minimum Mean Square Error}
  \acro{fnn}[FNN]{Feedforward Neural Network}
  \acro{eh}[EH]{Energy Harvesting}
  \acro{wpt}[WPT]{Wireless Power Transfer}
  \acro{dl}[DL]{Deep Learning}
  \acro{yolo}[YOLO]{You Only Look Once}
  \acro{mec}[MEC]{Mobile Edge Computing}
  \acro{modrl}[MODRL]{Multi-Objective Deep Reinforcement Learning }
  \acro{morl}[MORL]{Multi-Objective Reinforcement Learning}
  \acro{dwl}[DWL]{Deep W-Learning}
  \acro{per}[PER]{Prioritised Experience Replay}
  \acro{cnn}[CNN]{Convolutional Neural Network}
  \acro{dwn}[DWN]{Deep W-Networks}
\end{acronym}

\abstract{In this paper, we build on advances introduced by the Deep Q-Networks (DQN) approach to extend the multi-objective tabular Reinforcement Learning (RL) algorithm W-learning to large state spaces. W-learning algorithm can naturally solve the competition between multiple single policies in multi-objective environments. However, the tabular version does not scale well to environments with large state spaces. To address this issue, we replace underlying Q-tables with DQN, and propose an addition of W-Networks, as a replacement for tabular weights (W) representations. We evaluate the resulting Deep W-Networks (DWN) approach in two widely-accepted multi-objective RL benchmarks: deep sea treasure and multi-objective mountain car. We show that DWN solves the competition between multiple policies while outperforming the baseline in the form of a DQN solution. Additionally, we demonstrate that the proposed algorithm can find the Pareto front in both tested environments.}


\onecolumn \maketitle \normalsize \setcounter{footnote}{0} \vfill

\acresetall

\copyrightnotice

\section{Introduction}





Many real world problems such as radio resource management~\cite{giupponi2005novel}, infectious disease control~\cite{disease_control}, energy-balancing in sensor networks~\cite{hribar2020energy}, etc., can be formulated as a multi-objective optimisation problem. Whenever an agent is tackling such a problem in a dynamic environment, a single objective \ac{rl} methods such as Q-learning will not result in a behaviour that will be optimal for all objectives. Instead, the single objective solution will most likely prefer one objective over others. Alternatively, the agent can employ a \ac{morl} method. Unfortunately, while much work has been completed in tabular \ac{morl}~\cite{MultiObjectiveOverview}, the curse of dimensionality limits the applicability of these methods to real-world problems. The curse of dimensionality refers to the challenges that come with organizing and analyzing data that has an intractably and/or infinitely large state-space, for example, using images as input states. Recent developments in \ac{rl} merged Q-Learning with Neural Networks~\cite{AtariDeepMind}, vastly expanding the complexity of problems that could be tackled with \ac{rl}. Work has been done in the past few years in order to employ \acp{dqn} to solve Multi-Objective problems~\cite{MultiObjectiveOverview}. However, most of the proposed solutions have drawbacks. These drawbacks range from a required high number of sampled experiences to train the neural networks, which will take an extended amount of time, to adding complexity by creating new types of networks with altered memory storage. To overcome these obstacles, in this paper, we propose a deep learning extension to a tabular multi-objective technique called W-learning~\cite{WLearning}.

W-Learning was first proposed in the late 90's~\cite{WLearning}, as a multi-policy way to solve multi-Objective problems that use Q-Learning agents with a single objective as part of a larger system. The main principle is that there will be many Q-Learning agents, each with a different policy. These agents will all suggest an action that will be selfish, and the best action will need to be determined from these suggested selfish actions. The goal of W-Learning is to determine which of these actions should be selected, ensuring the right agent ``wins''. The way it determines this is by attempting to figure out how much the agent cares about the action that it suggests. Some scenarios or states might not impact the reward of certain agents so much however it may massively impact one or some of them. W-learning has been successfully applied in a range multi-objective problems, from speed limit control on highways~\cite{kusic2021spatial}, smart grid~\cite{Dusparic2015}, and conflict detection in self-adaptive systems~\cite{cardozo2020}.  However, as W-learning was proposed long before deep learning was successfully implemented in \ac{rl} to deal with the curse of dimensionality~\cite{AtariDeepMind}, this limited the domains and scale of problems that W-Learning can be applied to. In this paper, we take advantage of advances introduced by \ac{dqn} to train multiple \ac{ann} on different objectives and propose a new ``deep'' variant of W-Learning to address competition between these objectives. 

One of the most significant advantages of multi-policy algorithms over single-policy, e.g., Q-learning, is in their ability to find the \emph{Pareto Optimal} behaviour\cite{Pareto}. To be Pareto Optimal, the agent's actions must be such that an improvement in its decision process for an objective will not harm the reward for any other objective. Single-policy algorithms rely on some specification of preferences for given objectives and, therefore, will not necessarily find the policy that will result in the optimal Pareto front. In other words, multi-objective algorithms require less information about the environment before training and are generally more favoured for use in offline learning. 

Our proposed \ac{dwn} algorithm takes advantage of the computational efficiency of single-policy algorithms by considering each objective separately. These policies will suggest a selfish action that will only maximise their own reward. However, the \ac{dwn} resolves the competition between greedy single-objective policies by relying on W-values representing policies' value to the system. These W-values can be learned with interaction with the environment, following logical steps similar to a well-known Q-learning algorithm. In our proposed implementation, we employ two \acp{dqn} for each objective. One \ac{dqn} is used to learn a greedy policy for the given objective, while the second \ac{dqn} has only one output representing the policy's W-value for a given state input. Furthermore, \ac{dwn} has the benefit of training all policies simultaneously, which allows for a faster learning process. Additionally, \ac{dwn} can take advantage of modularity, meaning that policies can be trained separately and then included in the \ac{dwn} agent. Modularity also enables policies to be altered, e.g., the reward function is changed, added, or deactivated, without the need to re-train all other policies.

The rest of the paper is organised as follows. In the next section, we discuss the most important design features of \ac{drl} introduced in the last decade and related work. In section~3 we present and describe our proposed \ac{dwn} algorithm. Followed by evaluation section~4, in which we employ two multi-objective environments: multi-objective mountain car and deep sea treasure. We show in both environments that \ac{dwn} is capable of resolving the competition between multiple policies while outperforming the baseline in the form of \ac{dqn} solution. Finally, we provide concluding remarks in section~5.

\section{Background}

In this section, we introduce the essential elements required to understand \ac{drl} and review related algorithms capable of resolving multi-objective problems.

\subsection{Deep Reinforcement Learning}

The goal of \ac{rl} algorithm is to find the optimal policy $\pi_*$ for an environment that is fully characterised with an \ac{mdp} \cite{DeepLearningArticle}. With \ac{mdp} we describe a sequential decision-making process in a form of a state-space $\mathcal{S}$, an action space $\mathcal{A}$, a reward function $R$, and a set of transition probabilities between the states $\mathcal{P}$. 
In such settings, the optimal policy $\pi_*$ is the policy that will maximise the long-term reward. 



An example of a \ac{drl} algorithm that revolutionised the field in 2013 is \ac{dqn}. The \ac{dqn} algorithm is a deep learning extension of a well-known action-value algorithm named Q-learning.
In action-value based methods, the agent interacts with the environment by taking actions $a$ and receiving a reward $r$ that indicates if the taken action was desirable or not. The Q represents the quality of an action-value $Q(s, a)$, with $s$ representing the state. The objective of the \ac{rl} algorithm is to accurately estimate $Q$ values for all action-values using a Bellman equation. Once the agent can accurately determine all values, it can find the optimal policy  $\pi_*$ for selection actions that will maximise the expected reward $r + \gamma Q(s', a')$, with $\gamma$ representing the discount for rewards obtained in next time-step. The Q-values are updated in iterations as follows:

\begin{equation}\label{eq:update_q_values}
    Q_{i_+1}(s, a) = \mathbb{E}_{s \sim \mathcal{S}}\big[ r =\gamma \max Q^*(s',a')|s,a \big]; 
\end{equation}

\noindent and converge to the optimal value when:

\begin{equation}
    Q_i \rightarrow Q^* \phantom{A} \textrm{as}  \phantom{A}  i \rightarrow \infty.
\end{equation}

\noindent However, such an iterative approach requires the agent to explore the entire state space, i.e., try all possible action in every state. In practice, such an approach is impossible as such exploration would take a gargantuan amount of time and computational power.
Instead, a state-space approximator is used. An example of a very effective non-linear state-space approximator is \ac{ann}.




In the \ac{dqn}, a \ac{ann} function approximator with weights $\theta$ is employed to represent Q-network. The agent uses the network to estimate the action-values, i.e., $Q(s,a;\theta) \approx Q^*(s, a) $. The values in the \ac{ann} are updated, i.e., trained, at each iteration $i$ by minimising the loss:

\begin{equation}\label{eq:q_loss_function}
    L_i(\theta_i) = \mathbb{E}_{s, a\sim \rho (\cdot);s'\sim\mathcal{S}} \big [(y_i - Q(s,a;\theta_i))^2 \big],
\end{equation}

\noindent where $y_i = \mathbb{E}_{s\sim \mathcal{S}}[r + \gamma\max_{a'}Q(s',a';\theta_{i-1}|s,a]$ represent target and $\rho$ is the probability distribution over sequences. Note that when determining the target, the values from the previous iteration, i.e., $\theta_{i-1}$, are held fixed. The gradient of the loss function is then determined as:

\begin{multline}\label{eq:q_gradient_function}
\nabla_{\theta_i} L_i(\theta_i) =  \mathbb{E}_{s, a\sim \rho(\cdot);s'\sim\mathcal{S}} \big[\big( r +  \gamma\max_{a'}Q(s',a';\theta_{i-1} - \\ 
Q(s,a)\big) \nabla_{\theta_i} Q (s,a; Q_i) \big].
\end{multline}

\noindent In practice, the loss function is optimised using gradient descent as it is less computation-intensive than computing the expectation $\mathbb{E}$ directly. 
However, the training process can be unstable and prone to converge to a local optimum. To remedy this issue, the \ac{dqn} introduced experience replay and the use of policy and target \ac{ann}.





Experience replay is a batch memory $\mathcal{M}$ into which the agent is storing experiences. An experience typically consist of a state, the next state, selected action, and obtained reward, i.e., a tuple $ <s, a, s', r>$. The agent often samples experiences uniformly randomly during the training process. However, because not all experiences are equal in terms of importance to the learning process, a much better approach is to prioritise them, i.e., increase the probability of their selection. Using \ac{per}\cite{per} we determine the probability of sampling an experience $i$ as:

\begin{equation}\label{eq:per}
    \centering
    P(i) = \frac{p^\zeta_i}{\sum_K{p^\zeta_K}},
\end{equation}

\noindent with factor $\zeta, \zeta \in [0, 1]$ controlling the degree to which experiences are prioritised. Using \ac{per} can significantly reduce the  time the agent requires to find the optimal policy.
To further stabilise the learning process, the \ac{dqn} algorithm introduced the use of target \ac{ann} for estimating the $Q(s', a')$ during training. Such an approach is necessary because using the same \ac{ann} for determining both Q-values can results in very similar estimations due to possible small difference in $s$ and $s'$. Therefore, using a target policy for $Q(s', a')$ estimation can prevent such occurrences.  

In our work, we adopt these aforementioned advances to extend the original W-learning algorithm usability in large state spaces. 






\subsection{Multi-Objective Reinforcement Learning}

Existing \ac{rl} methods employed for resolving multi-objective optimisation problems can be generalised into two groups: tabular \ac{rl} and \ac{drl} methods. 

An example of a tabular methods are GM-Sarsa(0) \cite{GMSarsa} and its extension with weighted sum approach \cite{MOSinglePolicy}. GM-Sarsa(0) \cite{GMSarsa} aims to find good policies concerning the composite tasks as opposed to finding the best policy for each task and then merging these into a single policy. On the other hand, the authors in \cite{MOSinglePolicy} proposed to use a synthetic objective function to emphasise the importance of each objective in the form of weight for Q-values. Unfortunately, neither of these methods performs well in finding the optimal multi-objective policy. Additionally, the performance of the tabular methods deteriorates when applied in environments with large state spaces.  

The second group, \ac{drl} methods~\cite{SinglePolicyMODRL,TwoStageMODRL,modularSubsumption,dynamicWeights}, can deal with large state spaces. The deep optimistic linear support learning~\cite{SinglePolicyMODRL} is an example of the first known extension of the \ac{dqn} that dealt with multi-objectivity. The limitation of linear support approach is in redundant computations and additional required representations. Both limitations can be overcome in two-stage multi-objective \ac{drl}~\cite{TwoStageMODRL} approach. In the latter, once policies are learned, policy-independent parameters are tuned using a separate algorithm that attempts to estimate the Pareto frontier of the system. Similarly, modular multi-objective \ac{drl} with subsumption architecture~\cite{modularSubsumption} was proposed that combines the results of single policies, represented by a \ac{dqn}, to take the action most amenable to all rewards for each environment step. The approach resembles a voting system with Q-values representing a vote for a certain policy. Finally, dynamic weights in multi-objective \ac{drl}~\cite{dynamicWeights} were proposed to deal with situations where the relative importance of weights changes during training. In contrast, our proposed \ac{dwn} has the benefit of simultaneously training all policies, which allows for a faster dynamic adjustment of policy rewards that the above \acp{drl} methods lack.

\section{Deep W-Learning Framework}



\begin{figure*}
	\centering
	\includegraphics[width=5.5in]{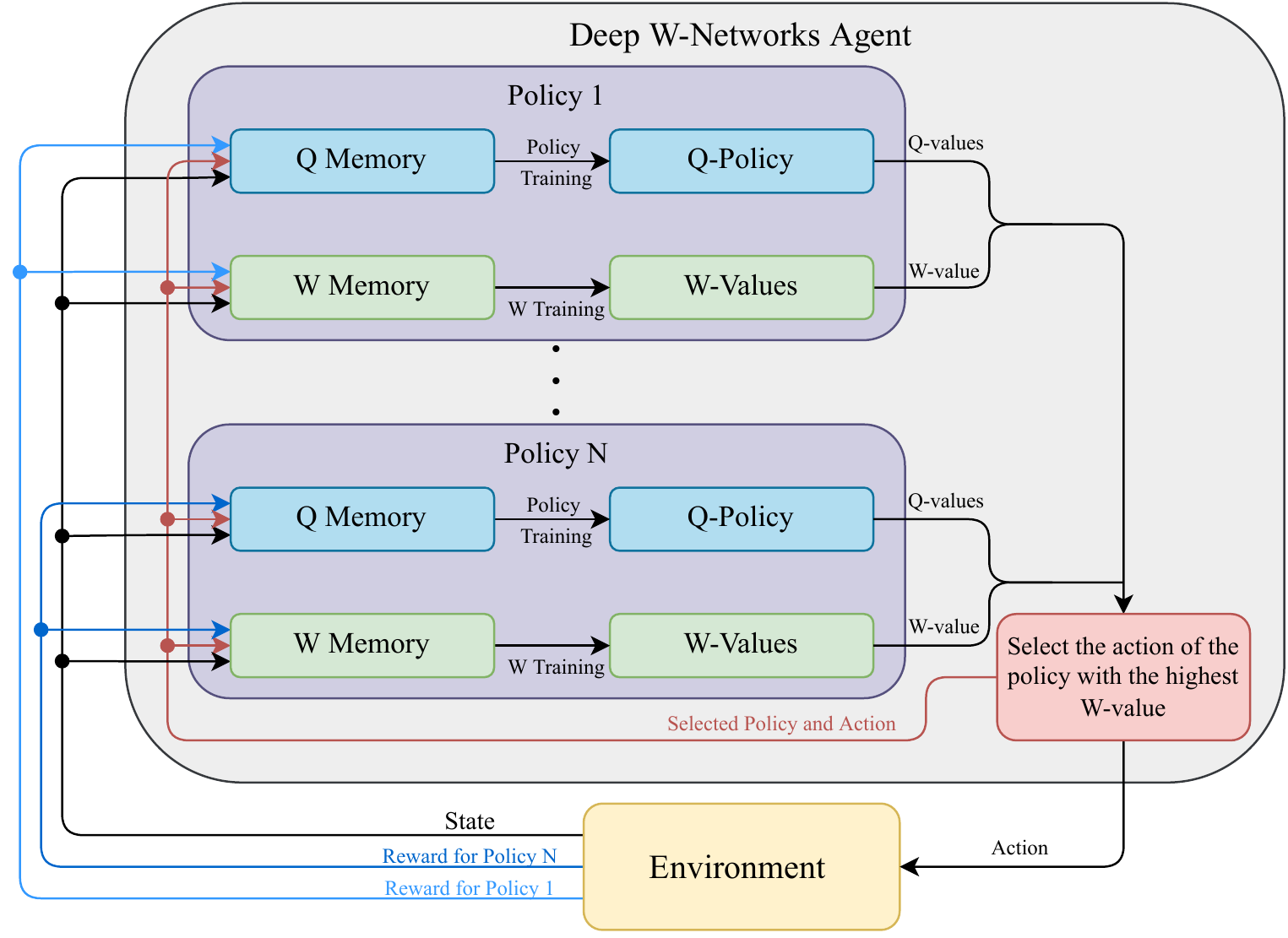}
	\caption{Deep W-Networks architecture. }
	\label{fig:DWN_overview}
	\vspace{-10pt}
\end{figure*}

In this section, we detail out our proposed multi-objective \ac{dwn} approach.  
We denote the multi-objective environment with tuple $ < N, \mathcal{S}, \mathcal{A}, \mathcal{R}, \Pi >$ in which $N$ is the number of policies, $\mathcal{S}$ represents the state space, $\mathcal{A}$ is the set of all available actions, $\mathcal{R} = \{R_1,...,R_N \}$ is set of reward functions,  and $\Pi = {\pi,...,\pi_N}$ denotes the set of available policies. At each decision epoch $t$ policies observe the same state $s(t) \in \mathcal{S}$ and every policy suggest an action. The agent's objective is to select the best possible action from the vector of actions nominated by agents for execution at that time-step $\mathbf{a}(t) = \{a_1(t),...,a_N(t)\}$. Furthermore, we denote the selected action at time-step $t$ with index $j$, i.e., the selected action is denoted as $a_j(t)$. Additionally, within our environment, we use $\Pi_{-j}$ to denote the set containing all polices except the $j$-th one.


The agent making the decision in a multi-objective environment at each time-step has to determine to what extent it will take into account each of the objectives. In other words, the agent has to continuously keep resolving the competition between multiple objectives. In W-learning, each objective is represented with a single Q-learning policy; each Q-learning policy has a different goal and, depending on the observed state $s(t)$, suggests a greedy action $a_i(t)$. These actions are often conflicting with each other, and to resolve the competition, the agent learns a table of weights for each state, called the W-values. For each observed state $s(t)$, the agent obtains $W_i(t)$ where $i$ is the index of the policy. The agent then takes the action suggested by the policy associated with the highest $W_i(t)$:

\begin{equation}\label{eq:w_max_values}
    W_j(t) = \max{ ( \{ W_1(t),...,W_N(t) \} )}.
\end{equation}

\noindent Note that the index $j$ marks the W-value of the policy the agent has selected. Updating the W-values follows a very similar formulation as updating the Q-values(Eq.\ref{eq:update_q_values}):

\begin{multline}\label{eq:update_w_values}
    W_i(t) \leftarrow (1-\alpha)W_i(t) + \alpha\big [Q(s(t), a_j(t)) - (R_i(t) + \\
    \gamma \max_{a_i(t+1) \in \mathcal{A}}Q(s(t+1), a_i(t+1)) \big].
\end{multline}

\noindent However, the agent will not update the W-value of the selected policy, i.e., the agent only updates W-values for set of policies $\Pi_{-j}$. Excluding W-Learning update for the selected policy allows other policies to emerge as the leader overtime. Such an approach is acceptable in practice as polices become more adept at their given task. Additionally, the updating constraint only applies for W-values, the agent will update Q-values for every policy, i.e., for the set $\Pi$.








\begin{algorithm}
\caption{Deep W-Learning with PER}\label{alg:dwl}
\begin{algorithmic}[1]
\State \textbf{Input:} Minibatch size $K$,  replay memory size $M$, exploration rates $\epsilon^Q$, $\epsilon^W$, smoothing factor $\zeta$, exponent $\beta$, w-learning learning rate $\alpha$, soft update factors $\tau^Q$, $\tau^W$
\State Initialize  $N$  Q-networks $\theta^Q_i$, target $\hat{\theta}^Q_i $, and replay memory $\mathcal{D}^Q_i$ $\forall i \in  \{1, \ldots, N\}$ 
\State Initialize  $N$  W-networks $\theta^W_i$, target $\hat{\theta}^W_i $, and replay memory $\mathcal{D}^W_i$ $\forall i \in  \{1, \ldots, N\}$ 
\For {$t=0$ to $T-1$}
\State Observe $s(t), s(t) \in \mathcal{S}$ 
\State // Nominate actions using epsilon-greedy 
\newline $\phantom{AAAA}$ approach ($\epsilon_Q$) 
\State Get $\mathbf{a}(t) = \{a_1(t),...,a_N(t)\}$
\State Get $ W_i(t) \forall i \in  \{1, \ldots, N\}$
\State // Select and execute action using 
\newline $\phantom{AAAA}$ epsilon-greedy approach ($\epsilon_W$)
\State Get $j$-th policy with the highest W-value 
\newline $\phantom{AAA}$ using Eq.~\ref{eq:w_max_values}
\State Execute action $a_j(t)$, and observe 
\newline $\phantom{AAA}$ state $s(t+1)$
\State Initiate policy training with Alg.~\ref{alg:qnetwork}
\State Initiate W training with Alg.~\ref{alg:wnetwork}
\EndFor
\end{algorithmic}
\end{algorithm}

In our proposed \ac{dwn} implementation, each policy has two \ac{dqn} networks.  The role of the first \ac{dqn} is to determine Q-values for the given policy, i.e., for greedy actions the policy will suggest. On the other hand, the second \ac{dqn} has only one output, representing the W-value, and replaces a tabular representation of state-W-value pairs present in the original W-learning implementation. We summarise the proposed \ac{dwn} in Alg.~\ref{alg:dwl}. Each policy requires two replay memories, one for Q-networks and another for the W-networks. Such a split is necessary because the agent will store a W-learning experience only when the agent did not select the policy.
Additionally, we employed \ac{per}~\cite{per} in our implementation to expedite the learning process.


\begin{algorithm}
\caption{Policy Training}\label{alg:qnetwork}
\begin{algorithmic}[1]
\State \textbf{Input}:$\Pi$, $\mathcal{D}^Q_i, i \in \{1,...,N\}$, $K$, $M$, $\zeta$, $\beta$, $\theta^Q_i, i \in \{1,...,N\}$, $\hat{\theta}^Q_i, i \in \{1,...,N\}$
\ForEach {$\pi_i$ in $\Pi$ }
\State Determine reward $r_i(t)$ using function $R_i$
\State Store experience $(s(t),a_j(t),r_i(t), s(t+1)$ 
\newline $\phantom{AA}$ in $\mathcal{D}^Q_i$  with priority $p_t =  \max_{m<t}p_m$
\ForEach{$k=1$ to $K$}
\State Sample transition $k \sim P(k)=p_k^\zeta / \sum_m p_m^\zeta$
\State Compute importance-sampling weight:
\newline $\phantom{AAAAA}$ $\omega_k = (M P(j))^{-\beta} / \max_m\omega_m$
\State Update transition priority $p_k \leftarrow | \delta_k|$
\State Accumulate weight-change: 
\newline $\phantom{AAAAA}$ $\Delta \leftarrow \Delta + \omega_k \cdot \delta_k \cdot \nabla_{\theta_i} Q(s(k-1), a_j(k-1)$
\EndFor
\State Update weights $\theta^Q_i \leftarrow \theta^Q_i + \eta \cdot \Delta $, reset $\Delta = 0$
\State Soft update target network:
\newline $\phantom{AAAAA}$ $\hat{\theta}^Q_i \leftarrow   \tau^Q \theta^Q_i + (1 - \tau^Q) \hat{\theta}^Q_i $

\State Update $\epsilon^{Q}$ using decay
\EndFor
\end{algorithmic}
\end{algorithm}


\begin{algorithm}
\caption{W Training}\label{alg:wnetwork}
\begin{algorithmic}[1]
\State \textbf{Input}:$\Pi$, $\mathcal{D}^W_i, i \in \{1,...,N\}$, $K$, $M$, $\zeta$, $\beta$, $\theta^W_i, i \in \{1,...,N\}$, $\hat{\theta}^W_i, i \in \{1,...,N\}$ 
\ForEach { $\pi_i$ in $\Pi_{-j}$ }
\State Determine reward $r_i(t)$ using function $R_i$
\State Store experience $(s(t),a_j(t),r_i(t), s(t+1)$ 
\newline $\phantom{AA}$ in $\mathcal{D}^W_i$ with priority $p_t =  \max_{m<t}p_m$
\EndFor

\ForEach { $\pi_i$ in $\Pi$ }
\ForEach{$k=1$ to $K$}
\State Sample transition $k \sim P(k)=p_k^\zeta / \sum_m p_m^\zeta$
\State Compute importance-sampling weight:
\newline $\phantom{AAAAAA}$ $\omega_k = (M P(j))^{-\beta} / \max_m\omega_m$
\State Update transition priority $p_k \leftarrow | \delta_k|$
\State Accumulate weight-change $\Delta$ using Eq. \ref{eq:update_w_values}
\EndFor
\State Update weights $\theta^W_i \leftarrow \theta^W_i + \eta \cdot \Delta $, reset $\Delta = 0$
\State Soft update target network:
\newline $\phantom{AAAAAA}$ $\hat{\theta}^W_i \leftarrow   \tau^W \theta^W_i + (1 - \tau^W) \hat{\theta}^W_i $
\State Update $\epsilon^{W}$ using decay
\EndFor
\end{algorithmic}
\end{algorithm}

The most significant aspect of the proposed \ac{dwn} algorithm is the action-nomination step. In the action nomination step, each policy suggests a greedy action with an epsilon probability that it will select a random action. Similarly, the agent will select the policy with the highest W-value but with epsilon probability, it might decide to explore, i.e., select the policy randomly. W-values exploration is necessary to avoid a single policy prevailing at the start of the learning due to high randomly initialized values and batch learning. Before the learning process can start, the agent requires a minimum of $K$ experiences stored in the replay memory to start the training process. 


We summarise the steps to train the policy and W-networks in two algorithms and give an overview of the \ac{dwn} architecture in Fig.~\ref{fig:DWN_overview}. During policy training (Alg.~\ref{alg:qnetwork}), each policy network optimises for the highest reward for its target. Note that this process remains unchanged from the original \ac{dqn} implementation, but is an integral part of the proposed \ac{dwn}.
After Q~networks have been through a few updates the W~training (Alg.~\ref{alg:wnetwork}) can begin. We achieve the delay by keeping the batch size for both policy and W training the same, or greater. The W~policy saves the experience only when it was not selected. In Alg.~\ref{alg:wnetwork}, line~3 we save the W~experiences of all policies but $j$-th, which was selected (in line 10, Alg.~\ref{alg:dwl}). Consequently, we achieve the delay in training the W networks. Epsilon greedy approach of selecting W-values ensures that the agent does not select the same W network in every step at the start of the training. In the next section, we demonstrate how \ac{dwn} performs in a multi-objective environment.





\section{Evaluation}


\begin{table*}[ht]
	\centering
	\caption{Hyperparameters for the Mountain Car Environment.}
	\vspace{5pt}
	\label{table_hyper_param_mountain_car}
	\begin{tabular}{l|l||l|l||l|l}
		\begin{tabular}[c]{@{}c@{}} Hyperparameter\end{tabular} & 
		\begin{tabular}[c]{@{}c@{}} Value \end{tabular} &
		\begin{tabular}[c]{@{}c@{}} Hyperparameter\end{tabular} & 
		\begin{tabular}[c]{@{}c@{}} Value \end{tabular} &
		\begin{tabular}[c]{@{}c@{}} Hyperparameter\end{tabular} & 
		\begin{tabular}[c]{@{}c@{}} Value \end{tabular} \\
		\hline
		
			\begin{tabular}[c]{@{}c@{}}  $\gamma$
		\end{tabular} & 
		\begin{tabular}[c]{@{}c@{}}   $0.99$ 
		\end{tabular} &
		\begin{tabular}[c]{@{}c@{}}   $\alpha$ 
		\end{tabular} & 
		\begin{tabular}[c]{@{}c@{}}   $1*10^{-3}$ 
		\end{tabular} &
		\begin{tabular}[c]{@{}c@{}}   $\beta$
		\end{tabular} & 
		\begin{tabular}[c]{@{}c@{}}   $0.4$ 
		\end{tabular} \\

		\begin{tabular}[c]{@{}c@{}}  $\epsilon^Q_{start}$  \end{tabular} & 
		\begin{tabular}[c]{@{}c@{}}   $0.95$ \end{tabular} &
		
		\begin{tabular}[c]{@{}c@{}}   $\epsilon^Q_{decay}$ \end{tabular} & 
		\begin{tabular}[c]{@{}c@{}}   $0.995$ 
		\end{tabular} &
		
		\begin{tabular}[c]{@{}c@{}}  $\epsilon^Q_{min}$  \end{tabular} & 
		\begin{tabular}[c]{@{}c@{}}   $0.1$ \end{tabular} \\

		\begin{tabular}[c]{@{}c@{}}   $\epsilon^W_{start}$ \end{tabular} & 
		\begin{tabular}[c]{@{}c@{}}   $0.99$ 
		\end{tabular} &
		
		\begin{tabular}[c]{@{}c@{}}  $\epsilon^W_{decay}$  \end{tabular} & 
		\begin{tabular}[c]{@{}c@{}}   $0.9995$ 
		\end{tabular} &
		
		\begin{tabular}[c]{@{}c@{}}   $\epsilon^W_{min}$ \end{tabular} & 
		\begin{tabular}[c]{@{}c@{}}   $0.1$ 
		\end{tabular} \\

		\begin{tabular}[c]{@{}c@{}}  $\zeta$  
		\end{tabular} & 
		\begin{tabular}[c]{@{}c@{}}   $0.6$ 
		\end{tabular} &

		\begin{tabular}[c]{@{}c@{}}  $\tau^Q$ 
		\end{tabular} & 
		\begin{tabular}[c]{@{}c@{}}   $1*10^{-3}$ 
		\end{tabular} &
		\begin{tabular}[c]{@{}c@{}}   $\tau^W$
		\end{tabular} & 
		\begin{tabular}[c]{@{}c@{}}   $1*10^{-3}$ 
		\end{tabular} \\

		\begin{tabular}[c]{@{}c@{}}  Batch size  $K$ \end{tabular} & 
		\begin{tabular}[c]{@{}c@{}}   $1024$ \end{tabular} &
		
		\begin{tabular}[c]{@{}c@{}}  Memory size $M$ \end{tabular} & 
		\begin{tabular}[c]{@{}c@{}}   $1*10^4$ 
		\end{tabular} &

		\begin{tabular}[c]{@{}c@{}}  Q Optimizer
		\end{tabular} & 
		\begin{tabular}[c]{@{}c@{}}   Adam 
		\end{tabular} \\
		
		\begin{tabular}[c]{@{}c@{}} W Optimizer  
		\end{tabular} & 
		\begin{tabular}[c]{@{}c@{}}   Adam 
		\end{tabular} &
		
		\begin{tabular}[c]{@{}c@{}}  Q learning rate
		\end{tabular} & 
		\begin{tabular}[c]{@{}c@{}}   $1*10^{-3}$  
		\end{tabular} &
		
		\begin{tabular}[c]{@{}c@{}} W learning rate
		\end{tabular} & 
		\begin{tabular}[c]{@{}c@{}}   $1*10^{-3}$  
		\end{tabular} \\

\end{tabular}
\end{table*}

In this section, we evaluate\footnote{\ac{dwn} algorithm implementation and evaluation code is available on \href{https://github.com/deepwlearning/deepwnetworks}{github.com/deepwlearning/deepwnetworks}.} the proposed \ac{dwn} using two multi-objective environments: multi-objective mountain car and deep sea treasure. The state space in the first environment is hand-crafted and is represented by only a two-input state vector. The two inputs are the car's position and velocity. The simplified case enables us to analyse \ac{dwn} performance in more detail. On the other hand, in the second environment, the deep sea treasure, we use visual inputs as states to demonstrate that \ac{dwn} performs well in environments with large state spaces.

\subsection{Multi-Objective Mountain Car}
The first environment, called \textit{Multi-Objective Mountain Car} presents a scenario where a car is stuck in the middle of a valley. The car must reach the top of the valley. However, the car does not have enough power to reach the top by driving directly forward. Instead, the agent has to learn to first move away from its objective, by reversing up the hill to gain momentum, in order to reach it. We use the environment, with minor modifications, as defined in~\cite{vamplew2011empirical}. The only alteration we made is the maximal number of steps allowed in an episode. We set the limit to 2000 because the goal of analysis in this environment is to gain a deeper understanding of \ac{dwn} performance. 

The environment has three different objectives: time penalty, backward acceleration penalty, and forward acceleration penalty. As the name of each policy suggests, the time policy gives a negative reward in every time step, except in the state when the agent reaches the top, the backward acceleration policy gives a negative reward when the agent is accelerating backward, and analogously, forward acceleration penalty applies for the forward policy. The agent has three available actions: accelerate forwards, accelerate backward, or do nothing. We design the \ac{dwn} agent with three policies, one for each objective. For simplicity, we use the same \ac{ann} structure for all policies and for both Q and W networks, i.e., $\theta^Q_i, \hat{\theta}^Q_i, \theta^W_i, \hat{\theta}^W_i \forall i$.
We use a feedforward \ac{ann} structure with two hidden layers, each with 128 neurons. On the output layer, to ensure better stability of learning, we employ a dueling network architecture \cite{dualQ}, with 256 neurons. We list hyper-parameters in Table~\ref{table_hyper_param_mountain_car}. 

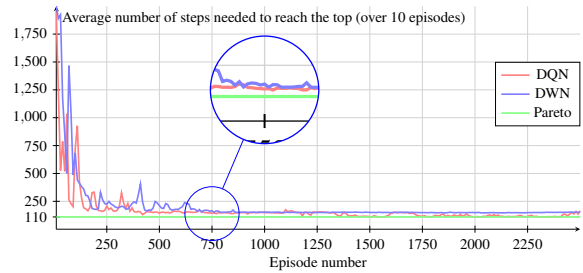
\begin{figure}
	\centering
	\begin{tikzpicture}[spy using outlines={circle, magnification=2, connect spies}]
\begin{axis}[
	axis lines=middle,
	xlabel=Episode number, 
	ylabel=Average number of steps needed to reach the top (over 10 episodes), 
	grid=both,
	minor grid style={gray!35},
	major grid style={gray!35},
	x label style={at={(axis description cs:0.5,-0.1)},anchor=north},
	ytick ={0, 110, 250, 500, 750, 1000, 1250, 1500, 1750},
	xtick ={25, 50, 75, 100, 125, 150, 175, 200, 225},
    xticklabels={250, 500, 750, 1000, 1250, 1500, 1750, 2000, 2250},
	width=5in,
    height=2.5in,
    font=\footnotesize,
	legend style={at={(1,0.75)}},
	ymin=0,ymax=2000,
	every tick/.style={
	        black,
	        semithick,
	}]
\addplot[line width=1pt,solid,color=red!50]%
	table[x=episodes,y=DQN_basic,col sep=comma]{csv_data/mountain_car_num_steps.csv};
	

\addplot[line width=1pt,solid,color=blue!50] %
table[x=episodes,y=DWL,col sep=comma]{csv_data/mountain_car_num_steps.csv};

\addplot[line width=1pt,solid,color=green!50]%
table[x=episodes,y=pareto,col sep=comma]{csv_data/mountain_car_num_steps.csv};

\addlegendentry{DQN};
\addlegendentry{DWN};
\addlegendentry{Pareto};

  \coordinate (spypoint) at (axis cs:75,140);
  \coordinate (magnifyglass) at (axis cs:100,1250);
\end{axis}

\spy [blue, size=2.3cm] on (spypoint)
   in node[fill=white] at (magnifyglass);

\end{tikzpicture}
	\caption{The number of steps, averaged over 10 episodes, each approach requires to finish the episode, i.e., the car reaching the top of the hill.}
	\label{fig:mountain_car_steps}
\end{figure}

\begin{figure}
	\centering
	\begin{tikzpicture}[spy using outlines={circle, magnification=2, connect spies}]
\begin{axis}[
	axis lines=middle,
	xlabel=Episode number, 
	ylabel=Policy selection in the episode (average over 10 episodes)($\%$),
	grid=both,
	minor grid style={gray!35},
	major grid style={gray!35},
	x label style={at={(axis description cs:0.5,-0.1)},anchor=north},
	ytick ={0,0.25,0.5, 0.75, 1},
    yticklabels={0,25,50, 75, 100},
	xtick ={25, 50, 75, 100, 125, 150, 175, 200, 225},
    xticklabels={250, 500, 750, 1000, 1250, 1500, 1750, 2000, 2250},
	width=5in,
    height=2.5in,
    font=\footnotesize,
	legend style={at={(1,0.6)}},
	ymin=0,ymax=1.1,
	every tick/.style={
	        black,
	        semithick,
	}]
\addplot[line width=1pt,solid,color=red!99]%
	table[x=episodes,y=policy1,col sep=comma]{csv_data/mountain_car_policy_selection.csv};
	

\addplot[line width=1pt,solid,color=blue!99] %
table[x=episodes,y=policy2,col sep=comma]{csv_data/mountain_car_policy_selection.csv};

\addplot[line width=1pt, solid,color=green!99] %
table[x=episodes,y=policy3,col sep=comma]{csv_data/mountain_car_policy_selection.csv};

\addlegendentry{Time penalty policy};
\addlegendentry{Backward acceleration penalty policy};
\addlegendentry{Forward acceleration penalty policy};

\end{axis}

\end{tikzpicture}
	\caption{The percentage each policy in DWN agent selects in an episode, averaged over 10 episodes.}
	\label{fig:mountain_car_policies1}
\end{figure}
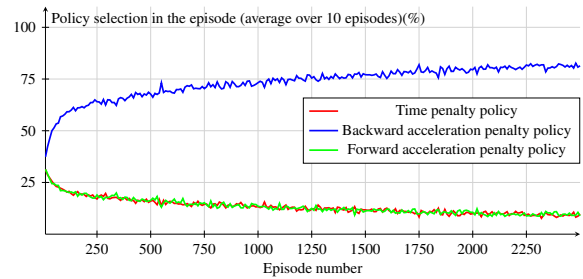

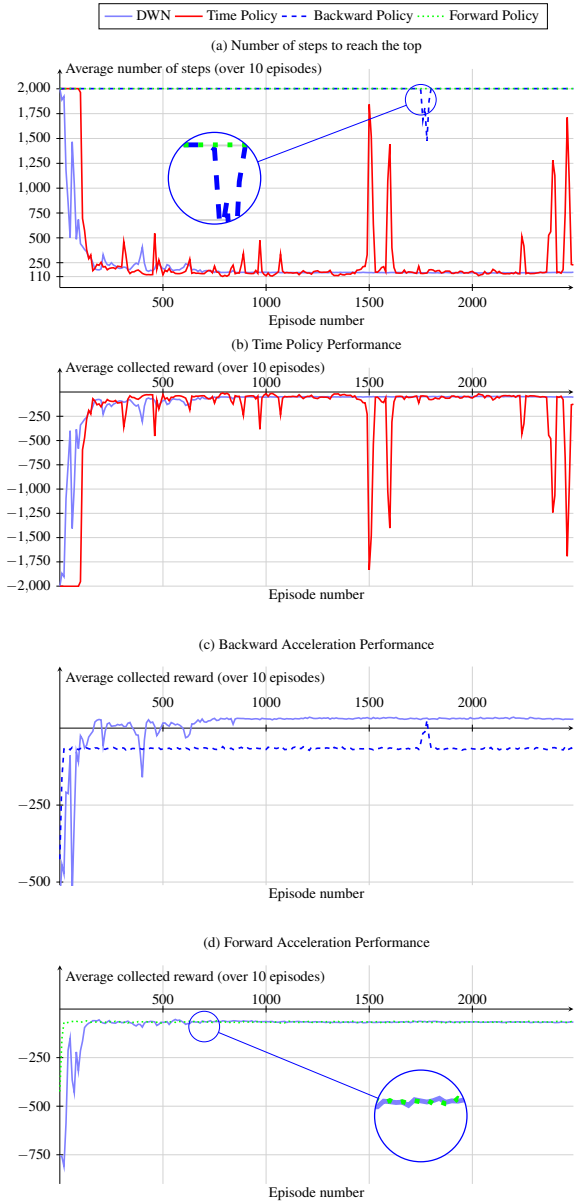
\begin{figure}
	\centering
	\begin{tikzpicture}[spy using outlines={circle, magnification=3, connect spies}]
\begin{groupplot}[group style={group name=my_plots, group size=1 by 4,horizontal sep= 1.25cm, vertical sep=1.75 cm}]]

\nextgroupplot[
    title = (a) Number of steps to reach the top,
	axis lines=middle,
	xlabel=Episode number, 
	ylabel=Average number of steps (over 10 episodes),
	ylabel style={yshift=0.25cm},
	grid=both,
	minor grid style={gray!35},
	major grid style={gray!35},
	x label style={at={(axis description cs:0.5,-0.1)},anchor=north},
	ytick ={0, 110, 250, 500, 750, 1000, 1250, 1500, 1750, 2000},
	xtick ={ 50, 100,  150,  200 },
    xticklabels={500, 1000, 1500, 2000},
	width=5in,
    height=2.5in,
    font=\footnotesize,
	legend style={at={(0.95,1.30)}, legend columns=-1},
	ymin=0,ymax=2200,
	every tick/.style={
	        black,
	        semithick,
	}]


\addplot[line width=1pt,solid,color=blue!50] %
table[x=i,y=DWL,col sep=comma]{csv_data/mountain_car_episode_results_steps.csv};


\addplot[line width=1pt,solid,color=red!99] %
table[x=i,y=time_policy,col sep=comma]{csv_data/mountain_car_episode_results_steps.csv};

\addplot[line width=1pt,dashed,color=blue!99] %
table[x=i,y=backward_policy,col sep=comma]{csv_data/mountain_car_episode_results_steps.csv};

\addplot[line width=1pt,dotted,color=green!99] %
table[x=i,y=forward_policy,col sep=comma]{csv_data/mountain_car_episode_results_steps.csv};

\coordinate (spypoint1) at (axis cs:175,1890);
        \coordinate (magnifyglass1) at (axis cs:75,1100);
        \spy [blue, size=2cm] on (spypoint1) in node[fill=white] at (magnifyglass1);

\addlegendentry{DWN};
\addlegendentry{Time Policy};
\addlegendentry{Backward Policy};
\addlegendentry{Forward Policy};

\nextgroupplot[
    title = (b) Time Policy Performance,
	axis lines=middle,
	xlabel=Episode number, 
	ylabel=Average collected reward (over 10 episodes), 
	ylabel style={yshift=0.25cm},
	grid=both,
	minor grid style={gray!35},
	major grid style={gray!35},
	x label style={at={(axis description cs:0.5,-0.1)},anchor=south},
	xticklabel shift=-15pt,
	ytick ={0, -250, -500, -750, -1000, -1250, -1500, -1750, -2000},
	xtick ={ 50, 100,  150,  200 },
    xticklabels={500, 1000, 1500, 2000},
	width=5in,
    height=2.5in,
    font=\footnotesize,
	ymin=-2002,ymax=250,
	every tick/.style={
	        black,
	        semithick,
	}]


\addplot[line width=1pt,solid,color=blue!50] %
table[x=i,y=DWL_time_rwd,col sep=comma]{csv_data/mountain_car_episode_results_rewards.csv};


\addplot[line width=1pt,solid,color=red!99] %
table[x=i,y=time_policy_rwd,col sep=comma]{csv_data/mountain_car_episode_results_rewards.csv};

\nextgroupplot[
    title = (c) Backward Acceleration Performance,
	axis lines=middle,
	xlabel=Episode number, 
	ylabel=Average collected reward (over 10 episodes), 
	grid=both,
	minor grid style={gray!35},
	major grid style={gray!35},
	x label style={at={(axis description cs:0.5,-0.1)},anchor=south},
	xticklabel shift=-25pt,
	ytick ={0, -250, -500},
	xtick ={ 50, 100,  150,  200 },
    xticklabels={500, 1000, 1500, 2000},
	width=5in,
    height=2.5in,
    font=\footnotesize,
	ymin=-510,ymax=200,
	every tick/.style={
	        black,
	        semithick,
	}]
	

\addplot[line width=1pt,solid,color=blue!50] %
table[x=i,y=DWL_back_rwd,col sep=comma]{csv_data/mountain_car_episode_results_rewards.csv};


\addplot[line width=1pt,dashed,color=blue!99] %
table[x=i,y=back_policy_rwd,col sep=comma]{csv_data/mountain_car_episode_results_rewards.csv};

\nextgroupplot[
    title = (d) Forward Acceleration Performance,
	axis lines=middle,
	xlabel=Episode number, 
	ylabel=Average collected reward (over 10 episodes), 
	grid=both,
	minor grid style={gray!35},
	major grid style={gray!35},
	x label style={at={(axis description cs:0.5,-0.1)},anchor=south},
	xticklabel shift=-15pt,
	ytick ={0, -250, -500, -750, -1000, -1250, -1500, -1750, -2000},
	xtick ={ 50, 100,  150,  200 },
    xticklabels={500, 1000, 1500, 2000},
	width=5in,
    height=2.5in,
    font=\footnotesize,
	ymin=-900,ymax=225,
	every tick/.style={
	        black,
	        semithick,
	}]


\addplot[line width=1pt,solid,color=blue!50] %
table[x=i,y=DWL_frwd_rwd,col sep=comma]{csv_data/mountain_car_episode_results_rewards.csv};


\addplot[line width=1pt,dotted,color=green!99] %
table[x=i,y=frwd_policy_rwd,col sep=comma]{csv_data/mountain_car_episode_results_rewards.csv};

\coordinate (spypoint4) at (axis cs:70,-90);
        \coordinate (magnifyglass4) at (axis cs:175,-550);
        \spy [blue, size=2cm] on (spypoint4) in node[fill=white] at (magnifyglass4);

\end{groupplot}
\end{tikzpicture}
	\caption{The comparison of performance between a policy as standalone DQN agent and a policy as a part of the proposed DWN.}
	\label{fig:mountain_car_policies2}
\end{figure}


In Fig.~\ref{fig:mountain_car_steps} we show the average number of steps, averaged over ten episodes, the agent requires to reach the top of the hill. The proposed \ac{dwn} and \ac{dqn} algorithms both achieve similar performance in the same number of episodes. Note that the \ac{dqn} receives the reward in the form of a sum of three reward signals, one for each policy the environment has. Furthermore, for a fair comparison \ac{dqn} employ the same \ac{ann} structure as our \ac{dwn}. Overall, the \ac{dwn} performance is similar, albeit slightly more stable, to \ac{dqn} in the mountain car environment. A far more interesting analysis is in individual policy performance by itself or part of \ac{dwn}.


In Fig.~\ref{fig:mountain_car_policies1} we show the percentage, i.e., how many times the 
\ac{dwn} agent has selected an individual policy in an episode. At the start, the agent selects policies evenly. Such a behaviour is expected due to high starting epsilon-greedy value. However, as the epsilon decays with the number of episodes and Q-learning \ac{dqn} learn, the agent starts to prefer one policy over the others. Interestingly, the backward accelerating policy proves to prevailing policy as the \ac{dwn} selects it three times more often the other two policies combined. 

In Fig.~\ref{fig:mountain_car_policies2} we compare \ac{dwn} with the performance of \ac{dqn} when it receives only the reward of a particular policy. In Fig.~\ref{fig:mountain_car_policies2} (a), we show the number of steps each policy requires to reach the top. Besides \ac{dwn} and the \ac{dqn} with time policy will reach the objective, i.e., arrive at the top of the hill. It appears, that when the reward signal is the only backward and forward policy the agent is unable, with a small exception, to learn to reach the top. However, when we look at the amount of collected reward a particular policy collects as part of \ac{dwn} or individually is almost the same. Meaning, that without exception policies learn to maximise their rewards. Note that the difference of $100$ in Fig.~\ref{fig:mountain_car_policies2} (c) between \ac{dwn} and backward acceleration policy is due to the reward signal. The agent receives a reward of $100$ when it reaches the top, and the backward policy reaches the top only as part of \ac{dwn} not individually. 

Combining the gained insights from the above results, we demonstrate that \ac{dwn} performs as expected: the agent is capable of reaching the end objective, while also maximise the reward collected by the individual policies within the \ac{dwn} agent.  In other words, policies in \ac{dwn} are selected in such a way that on an individual level each policy can achieve its best performance, i.e., maximise its long-term rewards. An added benefit is that the agent can also reach the main objective, i.e., reach the top of the hill.


\subsection{Deep Sea Treasure}

\begin{table*}[bt]
	\centering
	\caption{Hyperparameters for the Deep Sea Environment.}
	\vspace{5pt}
	\label{table_hyper_param_deep_sea}
	\begin{tabular}{l|l||l|l||l|l}
		\begin{tabular}[c]{@{}c@{}} Hyperparameter\end{tabular} & 
		\begin{tabular}[c]{@{}c@{}} Value \end{tabular} &
		\begin{tabular}[c]{@{}c@{}} Hyperparameter\end{tabular} & 
		\begin{tabular}[c]{@{}c@{}} Value \end{tabular} &
		\begin{tabular}[c]{@{}c@{}} Hyperparameter\end{tabular} & 
		\begin{tabular}[c]{@{}c@{}} Value \end{tabular} \\
		\hline
		
			\begin{tabular}[c]{@{}c@{}}  $\gamma$
		\end{tabular} & 
		\begin{tabular}[c]{@{}c@{}}   $0.9$ 
		\end{tabular} &
		\begin{tabular}[c]{@{}c@{}}   $\alpha$ 
		\end{tabular} & 
		\begin{tabular}[c]{@{}c@{}}   $1*10^{-3}$ 
		\end{tabular} &
		\begin{tabular}[c]{@{}c@{}}   $\beta$
		\end{tabular} & 
		\begin{tabular}[c]{@{}c@{}}   $0.4$ 
		\end{tabular} \\

		\begin{tabular}[c]{@{}c@{}}  $\epsilon^Q_{start}$  \end{tabular} & 
		\begin{tabular}[c]{@{}c@{}}   $0.95$ \end{tabular} &
		
		\begin{tabular}[c]{@{}c@{}}   $\epsilon^Q_{decay}$ \end{tabular} & 
		\begin{tabular}[c]{@{}c@{}}   $0.995$ 
		\end{tabular} &
		
		\begin{tabular}[c]{@{}c@{}}  $\epsilon^Q_{min}$  \end{tabular} & 
		\begin{tabular}[c]{@{}c@{}}   $0.25$ \end{tabular} \\

		\begin{tabular}[c]{@{}c@{}}   $\epsilon^W_{start}$ \end{tabular} & 
		\begin{tabular}[c]{@{}c@{}}   $0.99$ 
		\end{tabular} &
		
		\begin{tabular}[c]{@{}c@{}}  $\epsilon^W_{decay}$  \end{tabular} & 
		\begin{tabular}[c]{@{}c@{}}   $0.9995$ 
		\end{tabular} &
		
		\begin{tabular}[c]{@{}c@{}}   $\epsilon^W_{min}$ \end{tabular} & 
		\begin{tabular}[c]{@{}c@{}}   $0.01$ 
		\end{tabular} \\

		\begin{tabular}[c]{@{}c@{}}  $\zeta$  
		\end{tabular} & 
		\begin{tabular}[c]{@{}c@{}}   $0.6$ 
		\end{tabular} &

		\begin{tabular}[c]{@{}c@{}}  $\tau^Q$ 
		\end{tabular} & 
		\begin{tabular}[c]{@{}c@{}}   $1*10^{-3}$ 
		\end{tabular} &
		\begin{tabular}[c]{@{}c@{}}   $\tau^W$
		\end{tabular} & 
		\begin{tabular}[c]{@{}c@{}}   $1*10^{-3}$ 
		\end{tabular} \\

		\begin{tabular}[c]{@{}c@{}}  Batch size  $K$ \end{tabular} & 
		\begin{tabular}[c]{@{}c@{}}   $1024$ \end{tabular} &
		
		\begin{tabular}[c]{@{}c@{}}  Memory size $M$ \end{tabular} & 
		\begin{tabular}[c]{@{}c@{}}   $1*10^5$ 
		\end{tabular} &

		\begin{tabular}[c]{@{}c@{}}  Q Optimizer
		\end{tabular} & 
		\begin{tabular}[c]{@{}c@{}}   RMSprop 
		\end{tabular} \\
		
		\begin{tabular}[c]{@{}c@{}} W Optimizer  
		\end{tabular} & 
		\begin{tabular}[c]{@{}c@{}}   RMSprop 
		\end{tabular} &
		
		\begin{tabular}[c]{@{}c@{}}  Q learning rate
		\end{tabular} & 
		\begin{tabular}[c]{@{}c@{}}   $1*10^{-3}$  
		\end{tabular} &
		
		\begin{tabular}[c]{@{}c@{}} W learning rate
		\end{tabular} & 
		\begin{tabular}[c]{@{}c@{}}   $1*10^{-3}$  
		\end{tabular} \\

\end{tabular}
\end{table*}

The second environment, called \textit{Deep Sea Treasure}, is a simple grid-world with treasure chests that increase in value the deeper they are. The deeper the chest is, the further away from the agent it is. The goal of this scenario is for the agent to learn to optimise for future rewards rather than opting for the fractional short-term gain. We used the environment as proposed and implemented in~\cite{vamplew2011empirical}.

The deep sea environment has two objectives: time penalty and collected treasure. The time objective is for the agent to finish the episode, i.e., find the treasure, as quickly as possible. Therefore, the agent receives a negative reward of -1 at every step. The treasure reward depends on how deep is the treasure. Furthermore, the reward is increasing non-linearly with the depth and ranges from 1 to 124. In this scenario, our \ac{dwn} agent has two policies: time and treasure. As in the previous environment, all \ac{ann}, i.e., $\theta^Q_i, \hat{\theta}^Q_i, \theta^W_i, \hat{\theta}^W_i \forall i$, have the same \ac{cnn} structure. The first 2-dimensional convolution layer has three input channels and 16 output channels, and the second and third convolution layers have 32 channels. Every convolution layer has kernel size five with stride two, followed by batch normalisation. The last dense layer in the \ac{ann} has  1568 neurons. We list hyper-parameters in Table~\ref{table_hyper_param_deep_sea}.


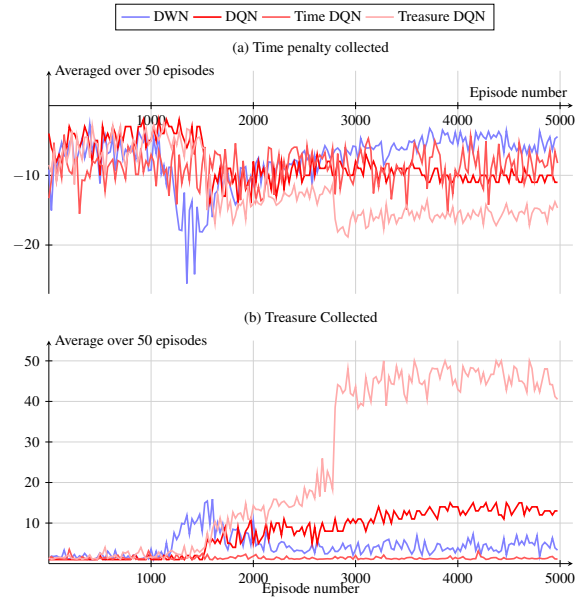
\begin{figure}
	\centering
	\begin{tikzpicture}
\begin{groupplot}[group style={group name=my_plots, group size=1 by 2,horizontal sep= 1.25cm, vertical sep=1.0 cm}]]

\nextgroupplot[
    title = (a) Time penalty collected,
	axis lines=middle,
	xlabel=Episode number, 
	ylabel=Averaged over 50 episodes,
	ylabel style={yshift=0.25cm},
	grid=both,
	minor grid style={gray!35},
	major grid style={gray!35},
	xtick ={40,  80, 120, 160, 200 },
    xticklabels={1000, 2000, 3000, 4000, 5000},
	width=5in,
    height=2.5in,
    font=\footnotesize,
	legend style={at={(0.85,1.30)}, legend columns=-1},
	xmin=0,xmax=205,
	ymin=-27, ymax=5,
	every tick/.style={
	        black,
	        semithick,
	}]


\addplot[line width=1pt,solid,color=blue!50] table[x=i,y=DWL_time,col sep=comma]{csv_data/treasure_episode_results.csv};

\addplot[line width=1pt,solid,color=red!99] table[x=i,y=DQN_time,col sep=comma]{csv_data/treasure_episode_results.csv};

\addplot[line width=1pt,solid,color=red!66] table[x=i,y=DQN_time_time,col sep=comma]{csv_data/treasure_episode_results.csv};

\addplot[line width=1pt,solid,color=red!33] table[x=i,y=DQN_tre_time,col sep=comma]{csv_data/treasure_episode_results.csv};

\addlegendentry{DWN};
\addlegendentry{DQN};
\addlegendentry{Time DQN};
\addlegendentry{Treasure DQN};

\nextgroupplot[
    title = (b) Treasure Collected,
	axis lines=middle,
	xlabel=Episode number, 
	ylabel=Average over 50 episodes, 
	ylabel style={yshift=0.25cm},
	xlabel style={yshift=-0.35cm},
	grid=both,
	minor grid style={gray!35},
	major grid style={gray!35},
	x label style={at={(axis description cs:0.5,-0.1)},anchor=south},
	ytick ={10, 20, 30, 40, 50},
	xtick ={ 40,  80, 120, 160, 200 },
    xticklabels={1000, 2000, 3000, 4000, 5000},
	width=5in,
    height=2.5in,
    font=\footnotesize,
	xmin=0,xmax=205,
	ymin=0,ymax=55,
	every tick/.style={
	        black,
	        semithick,
	}]

\addplot[line width=1pt,solid,color=blue!50] table[x=i,y=DWL_tre,col sep=comma]{csv_data/treasure_episode_results.csv};

\addplot[line width=1pt,solid,color=red!99] table[x=i,y=DQN_tre,col sep=comma]{csv_data/treasure_episode_results.csv};

\addplot[line width=1pt,solid,color=red!66] table[x=i,y=DQN_time_tre,col sep=comma]{csv_data/treasure_episode_results.csv};

\addplot[line width=1pt,solid,color=red!33] table[x=i,y=DQN_tre_tre,col sep=comma]{csv_data/treasure_episode_results.csv};

\end{groupplot}
\end{tikzpicture}
	\caption{Time penalty and treasure collected, averaged over 50 episodes, for the number of episodes.}
	\label{fig:deep_sea_policies}
\end{figure}

First, we analyse the performance of individual policies and compare it with \ac{dwn}. For the individual policy, we employed \ac{dqn} with the same neural structure as described above. In Fig.~\ref{fig:deep_sea_policies} we show the performance of three \ac{dqn} solutions, each with different reward signal. The first \ac{dqn} solution receives only the time penalty reward signal, the second only the treasure value signal, and the third the sum of the two reward signals. The  \ac{dqn} with only time reward learns to finish the episodes as fast as possible. Therefore, it learns to collect the first available treasure. The \ac{dqn} with only treasure reward learns to collect the highest treasure reward of all approaches. However, it does not learn to collect the highest treasure rewards, i.e., $74$ and $124$. The performance of \ac{dqn} with the sum of two rewards is exactly in the middle of the two. Interestingly, the \ac{dwn} performance is between the \ac{dqn} with the sum reward and \ac{dqn} with only time reward.


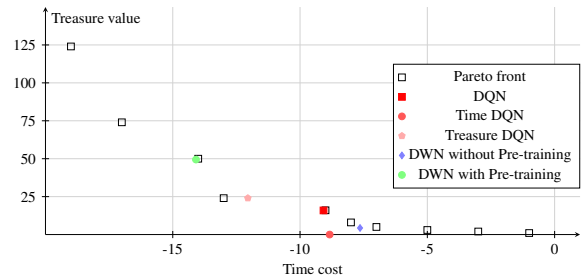
\begin{figure}
	\centering
	\begin{tikzpicture}
\begin{axis}[
	axis lines=middle,
	xlabel=Time cost,
	ylabel=Treasure value,
	grid=both,
	minor grid style={gray!35},
	major grid style={gray!35},
	x label style={at={(axis description cs:0.5,-0.1)},anchor=north},
	ytick ={25, 50, 75, 100, 125},
	xtick ={0, 5, 10, 15, 20},
    xticklabels={-20, -15, -10, -5, 0},
	width=5in,
    height=2.5in,
    font=\footnotesize,
	legend style={at={(1,0.75)}},
	xmin=0,xmax=21,
	ymin=0,ymax=150,
	every tick/.style={
	        black,
	        semithick,
	}]
\addplot[only marks, color=black, mark=square,]%
	table[x=time,y=treasure,col sep=comma]{csv_data/treasure_pareto.csv};
	
\addplot[only marks, color=red!99, mark=square*] %
    table[x=DQN_time,y=DQN_treasure,col sep=comma]{csv_data/treasure.csv};

\addplot[only marks, color=red!66, mark=*] %
    table[x=DQN_time_time,y=DQN_time_treasure,col sep=comma]{csv_data/treasure.csv};
    
    \addplot[only marks, color=red!33, mark=pentagon*] %
    table[x=DQN_tre_time,y=DQN_tre_treasure,col sep=comma]{csv_data/treasure.csv};

\addplot[only marks, color=blue!50, mark=diamond*] %
    table[x=DWL_pure_time,y=DWL_pure_treasure,col sep=comma]{csv_data/treasure.csv};
    
\addplot[only marks, color=green!50, mark=*] %
    table[x=DWL_time,y=DWL_treasure,col sep=comma]{csv_data/treasure.csv};

\addlegendentry{Pareto front};
\addlegendentry{DQN};
\addlegendentry{Time DQN};
\addlegendentry{Treasure DQN};
\addlegendentry{DWN without Pre-training };
\addlegendentry{DWN with Pre-training };

\end{axis}

\end{tikzpicture}
	\caption{The Pareto front for the deep sea treasure environment. }
	\label{fig:deep_sea_pareto}
\end{figure}

In Fig.~\ref{fig:deep_sea_pareto} we show how close to the Pareto front the agent with a different approach can arrive after 5000 episodes. Note that results are the average of the last one hundred episodes. The \ac{dqn} learns to reach the Pareto front. However, the collected treasure reward is far from ideal. The \ac{dwn}, when trained from scratch, is close but under-performs in comparison to a \ac{dqn} approach. Interestingly, \ac{dwn} with pre-trained Q-networks finds a high treasure at the Pareto front. In the latter approach, we took advantage of \ac{dwn} modularity properties. First, we trained time and treasure policy for 2500 episodes separately, and then for 2500 episodes, we trained as part of \ac{dwn}. Such a behaviour can be explained that policies as part of \ac{dwn} are not able to converge, thus giving them a head start, i.e., learning separately, can improve the \ac{dwn} performance. Furthermore, such a result was expected because, as it was pointed out in the original W-learning paper, we need to allow the Q-learning networks to learn first.

\section{Conclusion}


In this paper, we have proposed a deep learning extension to W-learning, an approach that naturally resolves competition in multi-objective scenarios. We have demonstrated the proposed method's efficiency and superiority to a baseline solution in two environments: deep sea treasure and multi-objective mountain car. In both of these environments, the proposed \ac{dwn} is capable of finding the Pareto front. Furthermore, we have also demonstrated the advantage of \ac{dwn} modularity properties by showing that using a pre-trained policy can aid in finding the Pareto front in the deep sea treasure environment. In our future work, we will focus on improving the computational performance and evaluating the performance in more complex environments, e.g., SuperMarioBros~\cite{gym-super-mario-bros}. 



The proposed \ac{dwn} algorithm can be employed in any system with multiple objectives such as traffic control, telecommunication networks, finance, etc. The condition being that each objective is represented with a different reward function. The main advantage of \ac{dwn} is its ability to train multiple policies simultaneously. 
Furthermore, sharing the state space between policies is not mandatory, e.g., a policy for the mountain car environment policies could only need access to the velocity vector. Meaning that with \ac{dwn} it is possible to train policies with different states due to the use of separate buffers for storing experiences.

\section*{\uppercase{Acknowledgements}}

This work was funded in part by the SFI-NSFC Partnership Programme Grant Number 17/NSFC/5224 and SFI under Frontiers for the Future project 21/FFP-A/8957.


\bibliographystyle{apalike}
{\small
\bibliography{example}}



\end{document}